\documentclass[conference]{IEEEtran}
\IEEEoverridecommandlockouts
\usepackage{cite}
\usepackage{amsmath,amssymb,amsfonts}
\usepackage{algorithm}
\usepackage{algpseudocode}
\usepackage{graphicx}
\usepackage{url}
\usepackage{hyperref}
\usepackage{textcomp}
\usepackage{xcolor}
\DeclareMathAlphabet{\mathpzc}{OT1}{pzc}{m}{it}
\usepackage{placeins}
\usepackage{multirow}
\usepackage{caption}
\usepackage{subcaption}
\usepackage{breqn}
\def\BibTeX{{\rm B\kern-.05em{\sc i\kern-.025em b}\kern-.08em
    T\kern-.1667em\lower.7ex\hbox{E}\kern-.125emX}}

\usepackage{color, colortbl}	
\definecolor{LightCyan}{rgb}{0.88,1,1}

\begin{document}



\title{Uncertainty-Autoencoder-Based Privacy and Utility Preserving Data Type Conscious Transformation\\
}
\author{\IEEEauthorblockN{\textsuperscript{}Bishwas Mandal}
\IEEEauthorblockA{\textit{Department of Computer Science} \\
\textit{Kansas State University}\\
 Manhattan, KS 66506, USA \\
 bishmdl76@ksu.edu}
\and
\IEEEauthorblockN{\textsuperscript{}George Amariucai}
 \IEEEauthorblockA{\textit{Department of Computer Science} \\
\textit{Kansas State University}\\
 Manhattan, KS 66506, USA \\
 amariucai@ksu.edu}
\and
\IEEEauthorblockN{\textsuperscript{}Shuangqing Wei}
\IEEEauthorblockA{\textit{Division of Electrical \& Computer Engineering} \\
\textit{Louisiana State University}\\
 Baton Rouge, LA 70803, USA \\
 swei@lsu.edu}
}

\maketitle

\begin{abstract}
We propose an adversarial learning framework that deals with the privacy-utility tradeoff problem under two types of conditions: data-type ignorant, and data-type aware. Under data-type aware conditions,  the privacy mechanism provides a one-hot encoding of categorical features, representing exactly one class, while under data-type ignorant conditions the categorical variables are represented by a collection of scores, one for each class. We use a neural network architecture consisting of a generator and a discriminator, where the generator consists of an encoder-decoder pair, and the discriminator consists of an adversary and a utility provider. Unlike previous research considering this kind of architecture, which leverages autoencoders (AEs) without introducing any randomness, or variational autoencoders (VAEs) based on learning latent representations which are then forced into a Gaussian assumption, our proposed technique introduces randomness and removes the Gaussian assumption restriction on the latent variables, only focusing on the end-to-end stochastic mapping of the input to privatized data. We test our framework on different datasets:  MNIST, FashionMNIST, UCI Adult, and US Census Demographic Data, providing a wide range of possible private and utility attributes. We use multiple adversaries simultaneously to test our privacy mechanism -- some trained from the ground truth data and some trained from the perturbed data generated by our privacy mechanism. Through comparative analysis, our results demonstrate better privacy and utility guarantees than the existing works under similar, data-type ignorant conditions, even when the latter are considered under their original restrictive single-adversary model.
\end{abstract}


\section{Introduction}
With the recent surge in the usage of online services, we share an increasingly large amount of data with different service providers in order to receive some form of utility. Even when we pay special attention to avoid disclosing what we deem to be private information, such as our identity, age, race, location, gender, income, medical conditions, political views etc., the data that we do share may still contain an uncomfortably large amount of information about these attributes -- an amount that may be just enough for intruders to infer our private/sensitive data. Recently, Google dropped its FLoC (Federated Learning of Cohorts) project after criticism that it could lead to sensitive data being inferred from user behavior and interests. It is therefore important to be aware of such potential data correlations, and to employ a privacy mechanism that can essentially destroy the information between the shared data and the private data, while preserving the useful information -- that required to achieve the desired level of utility. We call such a mechanism a \emph{privacy and utility preserving end-to-end transformation (PUPET)}.

Ideally, a PUPET should minimize the leakage of information about the private attributes, and maximize the information that the shared data contains about the utility attributes. The operation of a privacy mechanism is said to achieve a privacy-utility tradeoff when it establishes certain points in some private-information-utility-information plane. Most notions of privacy, such as Differential Privacy \cite{10.1007/11787006_1,10.1561/0400000042}, or k-anonymity \cite{Sweene02} are achieved by using some sort of distortion, such as adding random noise to the data, or performing data suppression or generalization. The privacy-utility tradeoff is the subject of a large body of research, with many works focused on producing application-specific solutions for the problem \cite{Rajagopalan2011, alvim2011differential, Makhdoumi2013PrivacyutilityTU, 5f7f56eb6ce4472fb78525d7b2181d5a, 9500410}. \cite{4529451} shows the drawbacks of k-anonymity and its variant. Similarly, it is known that computing the optimal noise addition in higher dimensional data for differential privacy can potentially be infeasible. Therefore, a wide majority of recent techniques use neural networks and adversarial learning, which can deal with the high dimensional data and can provide an approximation of the underlying functions. In particular, research works such as \cite{edwards2016censoring, e19120656, madras2018learning, pittaluga2018learning, 8515092, huang2019generative, chen2019distributed, wu2020privacypreserving, morales2020sensitivenets, Xiao_Tsai_Sohn_Chandraker_Yang_2020, erdemir2021active, wu2021privacypreserving} have been carried out leveraging autoencoder (AE) or variational autoencoder (VAE) \cite{kingma2014autoencoding} and/or generative adversarial network (GAN) type training \cite{goodfellow2014generative} or some other adversarial optimization techniques such as GRL \cite{ganin2015unsupervised} and K-Beam \cite{pmlr-v80-hamm18a}. Similarly, \cite{chen2018vganbased} use the loss function of both VAEs and GANs together along with their application specific classifier. AEs or VAEs are used to create compressed latent representations capturing minimum and maximum information about the private and utility features, respectively. AEs perform compression alone, without introducing randomness, while VAEs introduce randomness  when sampling from the mean and variance of the latent space -- they also model explicitly the prior distribution over the latent variable. VAE loss functions include a regularization term to minimize the KL divergence between the variational posterior over the latent variable and some prior distribution (which, for the convenience of generating data in the absence of an encoder, is usually chosen to be white Gaussian). GANs, on the other hand, involve a min-max game between the generator and discriminator. Similarly, recent obfuscation mechanisms \cite{RavalMachanavajjhalaPan, ilanchezian2019maximal, pmlr-v108-hsu20a} demonstrate privacy preservation by first estimating each feature's information density relative to the private data, and then selectively distorting the most information-leaking features. 

All these different techniques require either training a filter that adds noise, or adding noise directly into the input stream while training, or introducing distortion via lossy compression or some other form of masking technique. In addition, most of these techniques are tested on image datasets, for which there is no restriction on the output of data, compared to categorical features, where the output should be a one-hot encoding that represents precisely one class. In the adversarial learning framework, most research involving datasets that include categorical features either uses an adversary trained directly on the latent variables \cite{edwards2016censoring, chen2019distributed}, or one trained on the generator output but without enforcing any constraints based on the data type  (data-type-ignorant conditions) \cite{madras2018learning}. This lack of constraints mean that the output of the privacy mechanism doesn't represent an exact class, but rather some scores associated with different classes. The notion of adding noise or distortion to categorical variables in data publishing can be impractical as users are likely to publish data at a higher level where they will be limited to mentioning one particular class. Another notable issue in existing works is that most of the privacy mechanisms are tested under a single and particular type of adversary network which cannot demonstrate the \emph{actual privacy leakage} as there might be different adversary models capable of inferring private features with higher accuracy which can diminish the claims of particular privacy mechanisms. For example, \cite{Raval2017ProtectingVS} proposed the \emph{strong adversary} -- an adversary that is trained using the perturbed data generated by their privacy mechanism and ground truth labels -- and show that the adversary's accuracy increased from 50\% (weak adversary -- trained using ground truth image and labels) to 75\% i.e. a 25\% (absolute) increase in inferring private information. In order to make the privacy mechanism robust against multiple adversaries, \cite{wu2020privacypreserving, wu2021privacypreserving} propose \emph{privacy budget model restarting and ensemble}.

Inspired by these studies, in this paper, we propose an adversarial learning framework that deals with the privacy-utility tradeoff problem under two conditions: data-type-ignorant, and data-type-aware conditions. Data-type-ignorant refers to the classic privacy-utility tradeoff setting where the categorical variables after privatization are not enforced to belong to one particular class. The data-type-aware condition, on the other hand, refers to the optimization settings where the output of the \emph{PUPET} generator enforces the correct representation of categorical variables by assigning a one to the \emph{argmax} index of the probability distribution and assigning zero to all other indices before passing the data to the adversary and utility provider in the \emph{testing phase}. Thereafter, we test our privacy mechanism on multiple adversaries-- some trained from ground truth data and true labels (weak adversaries) and some trained from the perturbed data generated from the privacy mechanism and true labels (strong adversaries). Furthermore, in our proposed privacy mechanism, we introduce randomness and remove the Gaussian assumption on the prior distribution of the latent variable and only focus on the end-to-end stochastic mapping to transform data that preserves privacy and utility. The Gaussian assumption used by VAEs is useful for ancestral sampling, which is not required in the generation of privatized data.  Instead, we require a Markov chain $(X \xrightarrow{} Z \xrightarrow{} \hat{X})$ where $X$ is the input data, $Z$ is the latent variable and $\hat{X}$ is the privatized data. To implement our privacy mechanism, we leverage Uncertainty Autoencoders (UAEs) \cite{grover2019uncertainty}, which define an implicit generative model without specifying a prior on the latent representation. The use of a Markov chain to generate privatized data is necessary regardless of the generative model (AE, VAE, or UAE) and thus, the use of UAE doesn't make the process any more computationally expensive than previous approaches. In our adversarial setting, UAE behaves as a generator and learns from its own loss, along with the loss of the discriminators, to generate privatized data that maintains its utility to the utility provider. On the other hand, the discriminators (adversary and utility provider) learn to infer private and utility features, respectively, from the privatized data generated by the generator.

\begin{figure}[!htb]
\centerline{\includegraphics[width=0.5\textwidth]{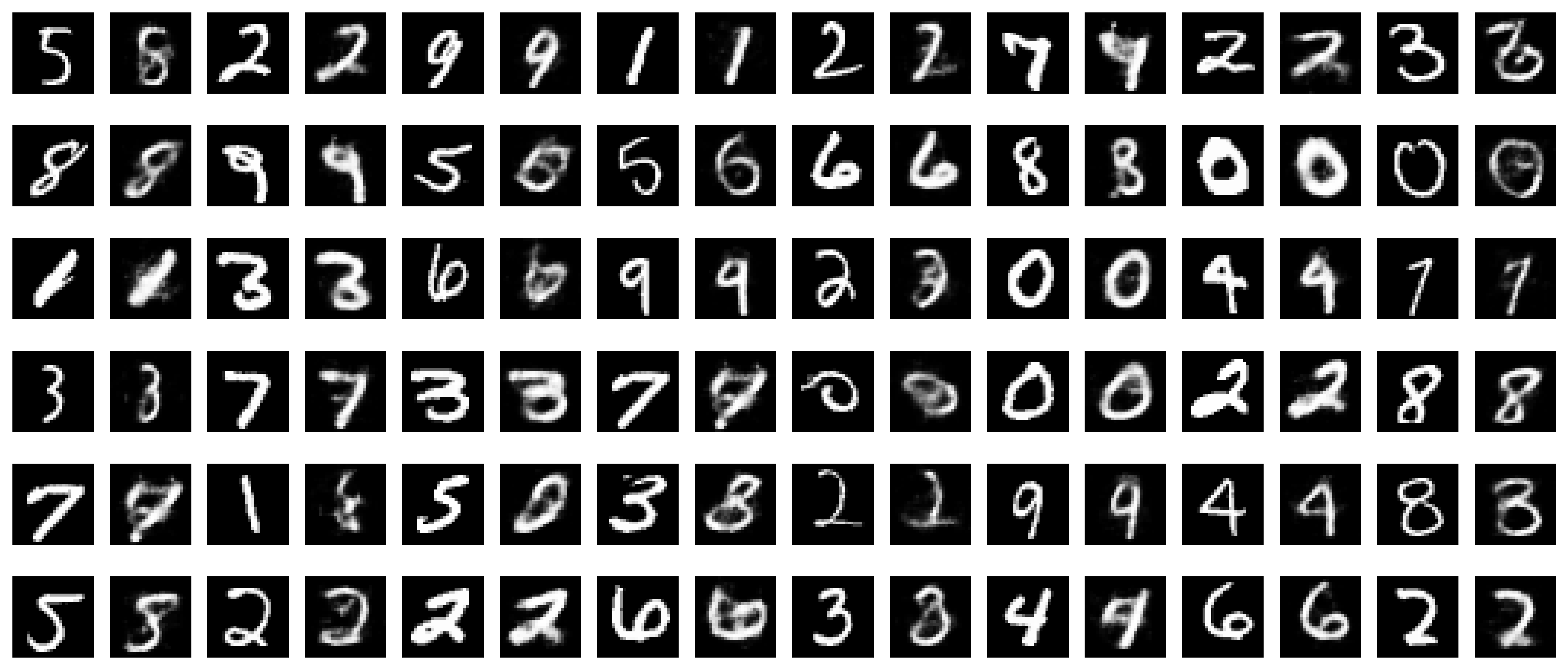}}
\caption{\textbf{MNIST}: Odd and even adjacent columns show original and privatized versions respectively. For most images, numbers are still in the same category (utility attribute: $\geq 5$ or $< 5$) while being switched from odd to even (private attribute). Some digits change from odd to even but also switch from $\geq 5$ to $< 5$, and some remain unchanged.}
\label{fig1}
\end{figure}

To showcase the effectiveness of our proposed privacy mechanism, we perform comprehensive experiments on four datasets viz. MNIST handwritten digits \cite{lecun-mnisthandwrittendigit-2010}, Fashion MNIST \cite{xiao2017fashionmnist}, UCI Adult \cite{Dua:2019}, and US Census Demographic Data \cite{uscensus}. Except UCI Adult, the other datasets do not contain  categorical features and thus fit well with data-type-ignorant conditions. Firstly, we evaluate our proposed method under data-type-ignorant conditions for all the four datasets considering multiple adversary models and compare results with the existing mechanisms. In particular, we directly compare our proposed UAE-based PUPET (or \emph{UAE-PUPET}) mechanism to that of \cite{chen2019distributed}, in the same setting as in \cite{chen2019distributed}, using their MNIST Case2 (variant). We show that our mechanism attains 2.6\% less accuracy for the private feature, and 11.2\% higher accuracy for the utility feature, thus clearly outperforming the previously existing mechanism. It  is  worth  mentioning  that  this result  was  obtained  by  our UAE-PUPET under the best-performing adversary out of multiple adversaries that includes weak  adversary,  strong adversary, and adversary trained together with utility provider and privacy mechanisms, at the time of adversarially training the privacy mechanism. This shows that our mechanism is fairly robust, despite the use of a single-adversary model at the time of training, instead of an ensemble of adversaries as in \cite{wu2020privacypreserving} and \cite{wu2021privacypreserving}, and also provides better privacy and utility guarantees than \cite{chen2019distributed}, which is tested against a single adversary model. Thereafter, we evaluate our privacy mechanism under data-type-aware conditions for the UCI Adult dataset and demonstrate similarity in performance between the claimed leakage (data-type-ignorant condition when dataset has categorical features) and actual leakage in practical applications.

The rest of the paper is organized as follows. The problem formulation and methodology is detailed in Section \ref{problemForm}, experimental results are given in Section \ref{experiment}, and concluding remarks are drawn in Section \ref{conclusions}.

\section{Problem Formulation and Methodology}\label{problemForm}
Consider a setting where a user wishes to release some data vector $X$ with the intent to receive certain level utility, while maintaining a certain level of privacy about a specific feature or set of features. We represent the private feature vector as $X_P$ and the utility feature vector as  $X_U$, and expect that they are both correlated with $X$. To ensure the desired privacy and utility guarantees, before publicly sharing their data $X$, users employ a PUPET that takes $X$ as input, and generates $\hat{X}$, a distorted version of $X$ which contains minimum information about $X_P$ and maximum information about $X_U$. The data $\hat{X}$ is then shared publicly.

\begin{figure} [htb!]
\centerline{\includegraphics[width=0.5\textwidth]{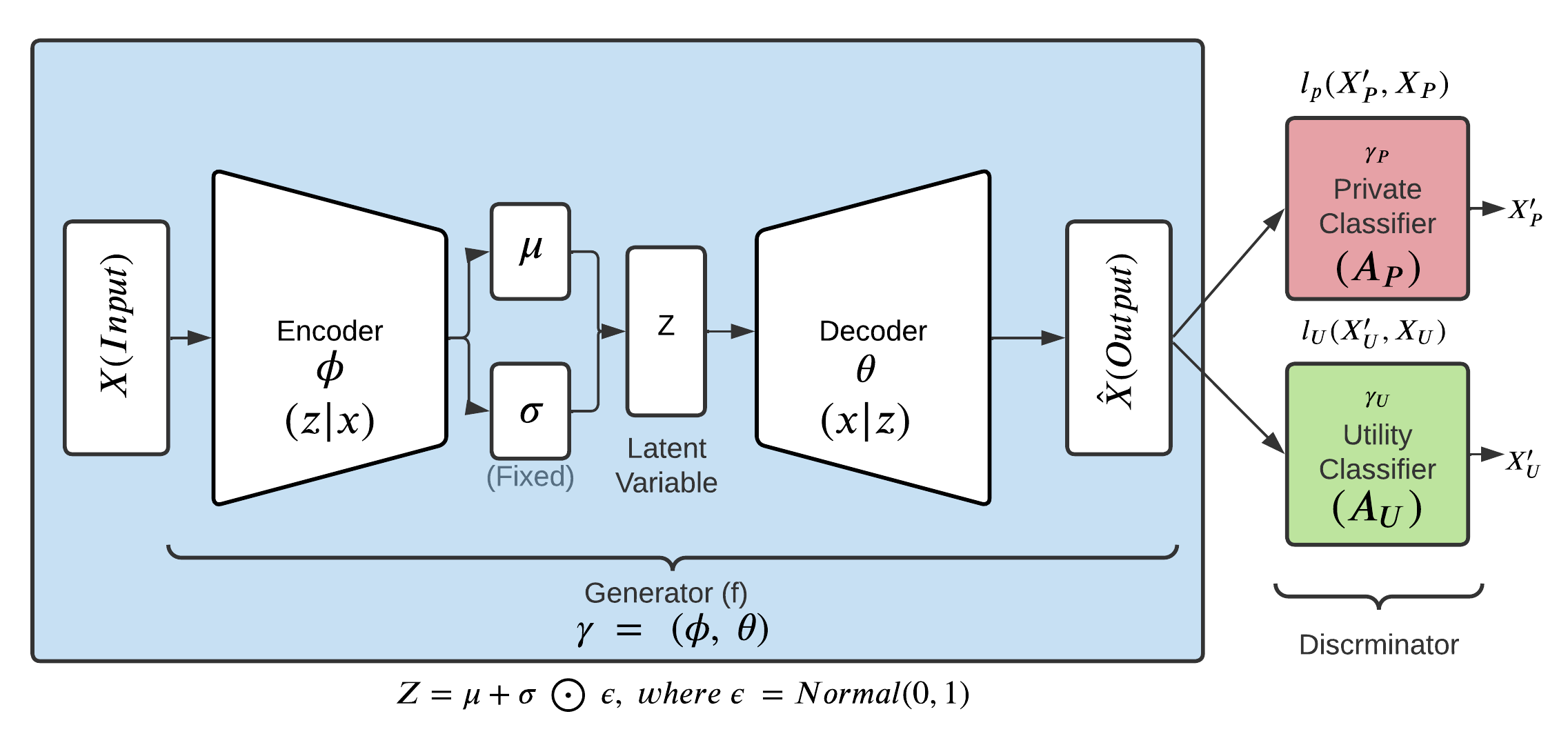}}
\caption{\textbf{UAE-PUPET} architecture. This architecture supports both conditions (data-type-ignorant and aware conditions). After the completion of training, we detach the discriminator, and use the generator to generate privatized data $\hat{X}$.}
\label{fig2}
\end{figure}

It is very important to note here that the disclosed data $\hat{X}$ is usually required to preserve in general the size and structure of $X$. That is, while it may be tempting to devise a compression mechanism -- for example, through a simple arbitrary affine transformation -- that produces a shorter $\hat{X}$, each component of which is some affine transformation of the components of $X$, such a mechanism has very little applicability in practice. This is because usually the disclosure of $\hat{X}$ has to take place over a pre-existing platform, outside of the data owner's control, which is designed specifically for $X$. This platform usually has very specific fields that the user needs to fill out, so that the structure of $X$ is enforced on the PUPET's output. It is also worth noting that most currently existing platforms (such as social media platforms) would provide their utility by taking the values of $\hat{X}$ at face value, as if it was the original $X$ that was being disclosed, which motivates some of the existing privacy-utility tradeoff works to use the distance between $\hat{X}$ and $X$ as a measure of utility \cite{asoodeh2015maximal, erdogdu2015privacy, wang2017estimation, kalantari2017information, basciftci2016privacy, wang2018utility, rassouli2019optimal, diaz2019robustness}. However, unlike these works, we consider a smart and informed utility provider, who is aware of the privacy mechanism employed by the user -- of course, without knowing the exact realizations of the randomness it uses, and can make a competent inference about the utility feature, using a neural-network-based architecture.

This setup raises questions as to who would use such services in real life, as well as whether such a setup would actually occur. One of the many examples is the creation of an e-commerce account, when certain information would need to be entered. The underlying platform gathers the information we share, and may develop implicit models to recognize, e.g., our gender and income. However, users might not be comfortable knowing that these implicit models learn about their income, but at the same time would also want to get better recommendation of products based on their gender. In such a case, the users can use a PUPET which generates $\hat{X}$, and then use $\hat{X}$ to fill up the information to create an account. The training of the  PUPET requires multiple tuples $(X, X_P, X_U)$, collected from users who do not mind disclosing $X_P,~ X_U$, and can be handled either by the data owner, or by a trusted service provider. Similarly, social media giants such as Facebook, Instagram, Twitter etc. sell users' data to different vendors. Such platforms can also use the formulated privatization services by ensuring the vendor only receives the data they paid for, i.e. privatize all the information about other features which they don't get paid for. Countless other applications can be envisioned.

\begin{algorithm}[!htb]
\caption{UAE-PUPET -- pseudocode for training and testing for data-type-ignorant and data-type-aware conditions}\label{alg:cap}
\begin{algorithmic}
\State // One step training
\State \textbf{Step 1:}
\State $\hat{\textbf{X}} \gets \textbf{f}(X, X_{P}, X_{U})$
\State \textbf{Step 2:}
\State $X'_{P} \gets a_{p}(\hat{\textbf{X}})$,then calculate cross entropy loss i.e. $l_P$
\State $X'_{U} \gets a_{u}(\hat{\textbf{X}})$, then calculate cross entropy loss i.e. $l_U$
\State \textbf{Step 3:}
\State Calculate generator loss and update $\gamma$.
\State // To calculate generator loss or $(gen\_loss)$:
\State calculate $\textit{reconstruction loss}$ i.e. mse$(X, \hat{\textbf{X}})$
\State $gen\_loss \gets \textit{reconstruction loss} + (\lambda_{U} * l_U) - (\lambda_{P} * l_P)$
\State Back propagate and update $\gamma$ parameters.
\State \textbf{Step 4:}
Take Step 1, then $X_{P}' \gets a_{p}(\hat{\textbf{X}})$, then calculate private_loss i.e. $l_P$
\State Back propagate and update $\gamma_{P}$ parameters.
\State \textbf{Step 5:}
Take Step 1, then $X_{U}' \gets a_{u}(\hat{\textbf{X}})$, then calculate utility_loss i.e. $l_U$
\State Back propagate and update $\gamma_{U}$ parameters.
\State // Testing
\State Take \textbf{Step 1} to generate privatized data $\hat{\textbf{X}}$.
\If{data-type-ignorant condition}
    \State pass // $\hat{\textbf{X}}$ without data type constraints
\ElsIf{data-type-aware condition}
    \For{each categorical variable ${\hat{\textbf{X}}[i:j}]$}
        \State $ \hat{\textbf{X}}[i:j] \gets enforce\_constraints({\hat{\textbf{X}}[i:j]})$
        \State // $enforce\_constraints$ assigns $1$ to argmax index
        \State // and assigns $0$ to other indices
    \EndFor
\EndIf
\State $X_{P}' \gets a_{p}(\hat{\textbf{X}})$, then evaluate private accuracy.
\State $X_{U}' \gets a_{u}(\hat{\textbf{X}})$, then evaluate utility accuracy.
\State // Recall privatized data is tested under multiple adversaries.
\end{algorithmic}
\end{algorithm}

Formally, denote  $X^j=\{x^j_1, x^j_2, x^j_3, \cdots x^j_{n^j}\}$, where the components $x^j_i$  are all correlated random variables denoting $n$ distinct features of the user $U^j$, which the user wishes to release to the public. In addition, the user $U^j$ contains private features $X^j_P$, and utility features $X^j_U$, with $n^j_p$ and $n^j_u$ component random variables, respectively. Note that $X^j_P$ and $X^j_U$ are both correlated with $X^j$ and no private and utility feature is in $X^j$ i.e. $X^j_P \notin X^j$ and $X^j_U \notin X^j$. For different users, the choice of the data they wish to share, as well as the private features and the utility features, differ and thus our privacy mechanism needs to be trained differently for different sets of users. For simplicity we drop the user-specific indices and refer to the random vectors directly as  $X$, $X_P$ and $X_U$. We represent the PUPET in its most general form as a function $f$ (which could be a randomized mapping) that takes input $(X, X_P, X_U)$ and generates $\hat{X}$ i.e. $\hat{X} = f(X, X_P, X_U)$. The adversary builds a learning algorithm $a_p$ that takes privatized data $\hat{X}$ to infer the private attributes $X_{P}'$ which is an estimate of $X_{P}$ i.e. $X_{P}' = a_p (\hat{X})$. The goal of adversary is to minimize the loss between $X_{P}'$ and $X_P$ i.e. $l_P(X_{P}', X_P) = l_P(a_p(f(X,X_P,X_U)),X_P)$. Correspondingly, the utility provider builds a learning algorithm $a_u$ to infer $X_{U}'$ which is an estimate of $X_U$ and desires to minimize the loss $l_U(X_{U}',X_{U}) = l_U(a_u(f(X,X_P,X_U)),X_U)$. The privacy mechanism $f$ is now chosen to maximize the inference loss $l_P$ and minimize the inference loss $l_U$. This setting refers to the min-max game expressed as follows:
\begin{equation}\label{eqn1}
\begin{split}
   \max_{f \in \mathpzc{F}}
   \Biggl\{ \lambda_{P}
   \min_{a_p \in \mathpzc{A_P}} \mathbb{E} \left[ l_{P} \left( a_p \left( f \left( X, X_P, X_U \right) \right) , X_P \right) \right] \\ - 
   \lambda_{U} \min_{a_u \in \mathpzc{A_U}} \mathbb{E} \left[ l_{U} \left( a_u \left( f \left( X, X_P, X_U \right) \right) , X_U \right) \right]
   \Biggr\},
\end{split}
\end{equation}where $\lambda_{P}, \lambda_{U}$ are hyperparameters that control the tradeoff between adversary loss and utility provider loss, and the expectation is taken over all samples of the dataset, and $\mathpzc{F}$, $\mathpzc{A_P}$ and $\mathpzc{A_U}$ are the sets of functions from which the privacy mechanism, the adversary and the utility provider select their corresponding operators, respectively. In this paper we shall use neural networks to optimize over different functions $f, a_p, a_u$ and the losses $l_{P}$ and $l_{U}$ are instantiated as the cross-entropy loss \cite{Boer04atutorial}.

In order to solve the optimization problem in Eq.\eqref{eqn1}, we leverage an uncertainty autoencoder (UAE), which serves as the generator for our privacy mechanism. The objective of UAE is given by $\max_{\theta,\phi} \mathbb{E}_{Q_{\phi}(X,Z)} \left[ \log p_{\theta}(x|z) \right]$, where $X$ is the input data distribution, $Z$ is the latent variable, $\phi, \theta$ are parameters of encoder and decoder, respectively, $Q_{\phi}(X,Z)$ is the true joint distribution of the input and latent representation, and $p_{\theta}(X|Z)$ is the likelihood function  produced by the decoder, which aims to emulate $Q_{\phi}(X|Z)$ as closely as possible. Notice that we don't force a Gaussian assumption, or any other distribution assumption, on the latent variable marginal distribution, and thus we don't have a KL divergence term in the objective of the UAE, as is the case for the VAE. Instead we focus only on the end-to-end stochastic mapping. Additionally, UAE's encoder outputs only the mean, and sampling is done in the latent space with a fixed variance. If the variance of the noise sampled in the latent space equals zero, the learning objective of the UAE is the same as that of a standard AE. In our adversarial setting, we introduce the parameter $\gamma$, which represents the weights of the generator (encoder-decoder pair) with function $f$ (basically, $\gamma=(\phi, \theta)$). Additionally, the output of the generator is attached to the discriminator, which consists of an adversary and a utility provider as shown in Fig. \ref{fig2}.
The adversary learns the function $a_p$ with parameters $\gamma_{P}$ to minimize the privacy-specific loss $l_P(X_{P}',X_{P})$. Similarly, the utility provider learns the $a_u$ with parameters $\gamma_{U}$ to minimize the utility-specific loss $l_U(X_{U}',X_U)$. Conversely, the generator function $f$ with parameter $\gamma$ learns to maximize $l_P(X_{P}',X_{P})$ and minimize $l_U(X_{U}',X_U)$ respectively along with the objective of UAE. In this setting, the neural network parameters $\gamma, \gamma_{P}, \gamma_{U}$ all are trained together to solve the following optimization problem shown in Eq.\eqref{eqn2}.
\begin{equation}\label{eqn2}
\begin{split}
    \max_{\gamma} \Biggl\{  \mathbb{E}_{Q_{\phi}(X,Z)} \left[ \log p_{\theta}(x|z) \right]  + \lambda_{P} \min_{\gamma_{P}} \left\{\mathbb{E}  \left[l_{P}\left(X_{P}',X_P\right)\right] \right\}  - \\ \lambda_{U}\min_{\gamma_{U}} \left\{\mathbb{E} \left[l_{U} \left(X_{U}',X_{U}\right) \right] \right\} \Biggr\} \text{, where}\\ X_{P}' = a_{p}(f(X,X_{P},X_{U})) \text{ , } X_{U}' = a_{u}(f(X,X_{P},X_{U})).
\end{split}
\end{equation}

\section{Experiments and Results}\label{experiment}
We perform comprehensive experiments on four widely-used datasets such as MNIST, Fashion MNIST, UCI-adult, and US Census Demographic Data, to demonstrate the effectiveness of our privacy mechanism i.e. solving the optimization problem in Eq.\eqref{eqn2}. Firstly, we discuss the results for data-type-ignorant conditions and compare them to existing works wherever possible and then present the result of data-type-aware conditions. We perform experiments with hyperparameter $\lambda_{P}$ = [0,10,20,...,100] for data-type-ignorant conditions and plot a UPT (Utility Privacy Tradeoff) curve. The UPT curve is used to represent the upper bound on the performance of the privacy-utility mechanism. As such, it consists of the upper convex hull of all the achieved operational points. The operational interpretation of the upper convex hull relies on an operational interpretation of any line connecting two operational points. Recall that each operational point is defined by two accuracy levels, achieved by averaging over an entire test dataset. If we split the test dataset in two parts of sizes $\alpha$ and $1-\alpha$ times the original size, respectively, and apply the first part to the mechanism achieving operational point $P_1$, and the second part to the mechanism achieving $P_2$, then the average over the entire dataset should achieve operational point $\alpha P_1 +(1-\alpha) P_2$. It is in this sense that the upper convex hull is achievable. Notice that for each $\lambda_{P}$ value, we conduct 25 experiments with random weight initialization and then report the mean for that particular $\lambda_{P}$. Similarly, the \emph{best guess point} refers to a certain accuracy that can be achieved by guessing a class which has the highest frequency. The pseudocode for our implementation is shown in Algorithm \ref{alg:cap}. For more information about the architecture of the generators, discriminators, and different hyperparameters used for multiple experiments, please refer to our source code in the following repository: \emph{\textcolor{blue}{\url{https://github.com/bishwasmandal246/uae-pupet}}}

\subsection{MNIST}
This experiment considers the same setting as \cite{chen2019distributed} and \cite{rezaei2018application} where
the private attribute refers to whether the number is odd or even, while the utility attribute encodes whether the number is $\geq 5$ or not. Fig. \ref{fig1}, shows the original and privatized images generated by our privacy mechanism.We also implemented our privacy mechanism using  autoencoders (AE-PUPET), variational autoencoders (VAE-PUPET), and $\beta$VAE \cite{Higgins2017betaVAELB} ($\beta$VAE-PUPET) to demonstrate empirically that removing the Gaussian restrictions from the latent variables (UAE-PUPET) is helpful in achieving better privacy and utility guarantee i.e. low private attributes accuracy and high utility attributes accuracy. Note that $\beta$VAE is an extension of VAE with $\beta > 1$ as the multiplier to the KL divergence term of VAE's learning objective. Table \ref{mnist-table} illustrates the upper bound accuracy, and area-under receiver operating characteristic curve (AUROC) results of our privacy mechanism \emph{UAE-PUPET}, and its comparison to emb-g-filter of \cite{chen2019distributed}, AE-PUPET, VAE-PUPET and $\beta$VAE-PUPET. The results for  \cite{chen2019distributed} are extracted from the paper, where the authors only provide accuracy scores, and hence AUROC is kept blank. Our proposed mechanism clearly outperforms emb-g-filter by achieving 2.6\% less (absolute) accuracy on the private attribute and 11.2\% higher (absolute) accuracy on the utility attribute even when UAE-PUPET is tested under multiple adversaries (weak adversary, strong adversary, and the adversaries trained during the privacy mechanism training)  whereas the emb-g-filter of \cite{chen2019distributed} is tested only under their original restrictive single adversary model. Similarly, with our setup, it is clear to see the use of UAE also provides better privacy and utility guarantees than that of AEs, VAEs or $\beta$VAEs. Fig. \ref{fig:upt-mnist} shows the UPT curve and compares different versions of PUPET. Each point in the curve represents the mean of 25 experiments for a particular $\lambda_{P}$ value. Note that the points in the upper-left (North-West) regions are the most favorable.

\begin{table}[!htb]
\resizebox{\columnwidth}{!}{
\begin{tabular}{|c|c|c|c|c|}
\hline
\multirow{2}{*}{Models} & \multicolumn{2}{c|}{Private attr.} 
&
\multicolumn{2}{c|}{Utility attr.} \\
\cline{2-5}
 & accuracy & AUROC & accuracy &  AUROC \\
\hline
without distortion(raw) & 0.98 & 0.98 & 0.98 & 0.98\\ 
emb-g-filter \cite{chen2019distributed} & 0.651 & - & 0.855 & -\\
VAE-PUPET & 0.620  & 0.619 & 0.836 & 0.834\\
$\beta$VAE-PUPET & 0.6117  & 0.6119 & 0.8032 & 0.81\\
AE-PUPET &  0.654 & 0.657 & 0.964  & 0.964 \\
\rowcolor{LightCyan}
UAE-PUPET & 0.625  & 0.622 & 0.967 & 0.967\\
\hline
\end{tabular}
}
\caption{\small MNIST accuracy and auroc results comparison}
\label{mnist-table}
\end{table}

\begin{figure}[!htb]
\minipage{0.244\textwidth}
  \includegraphics[width = \textwidth]{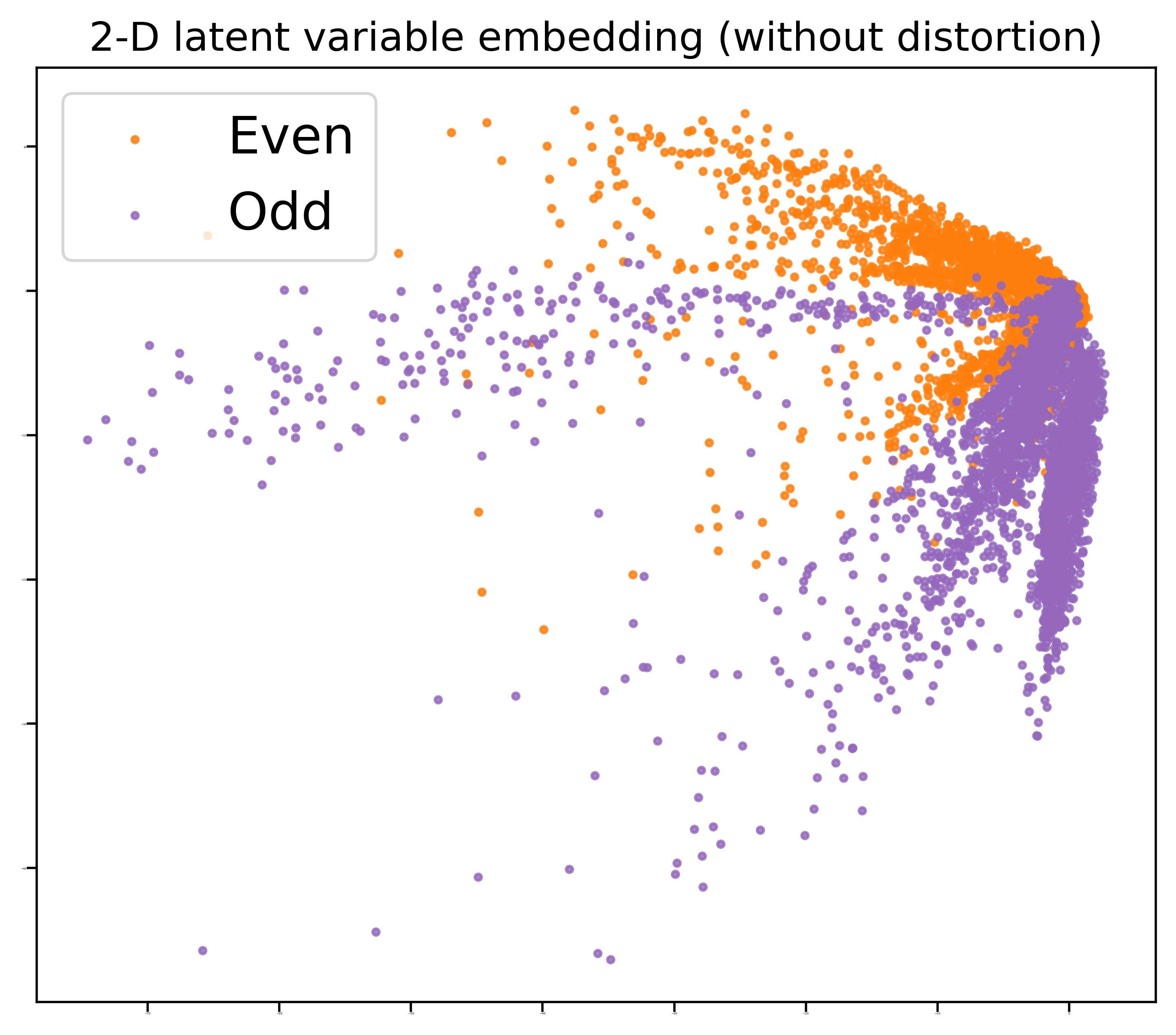}
    \subcaption{before UAE-PUPET (without distortion)}
    \label{fig:latent 1}
\endminipage
\minipage{0.244\textwidth}
  \includegraphics[width = \textwidth]{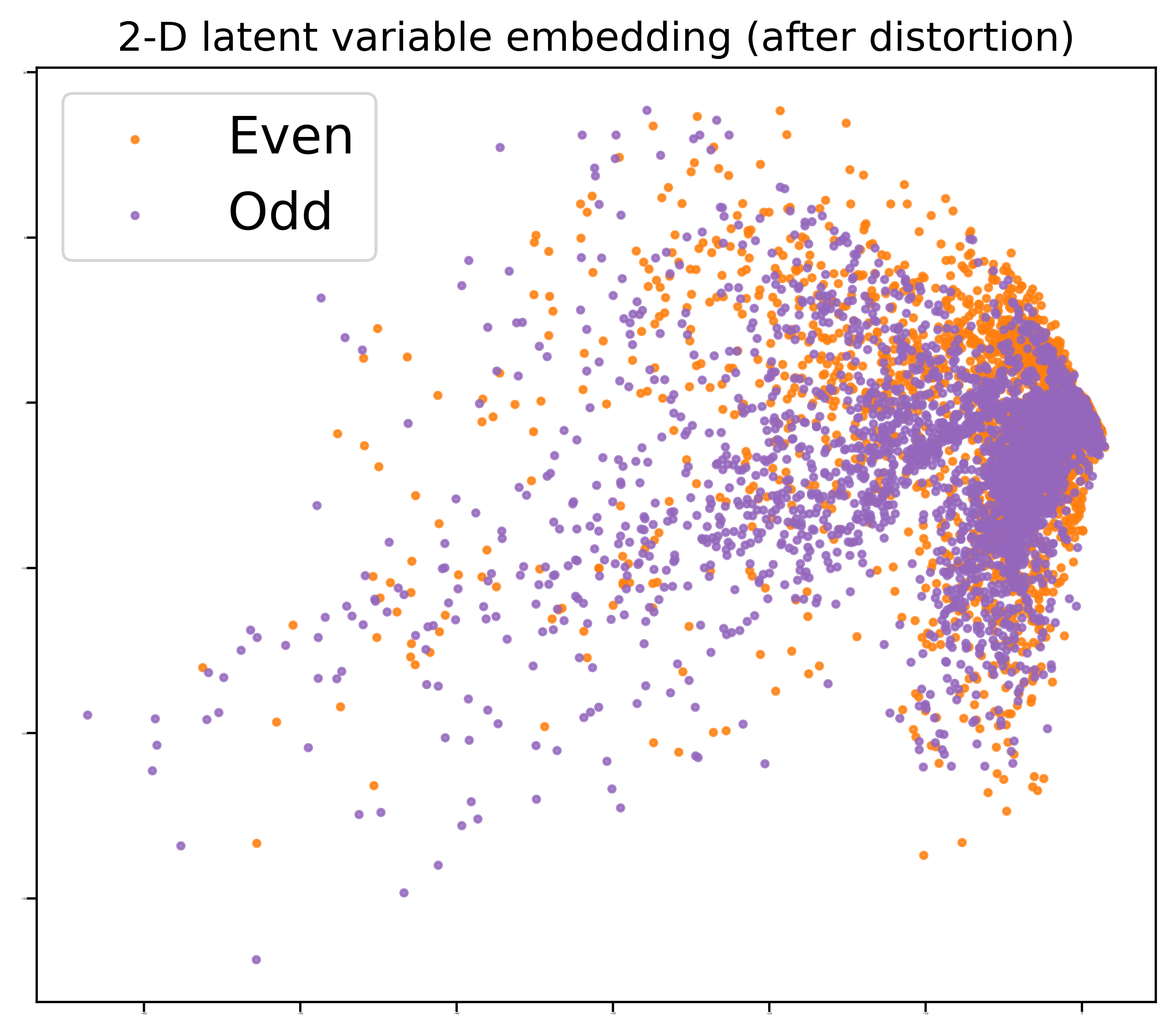}
    \subcaption{after UAE-PUPET}
    \label{fig: latent2}
    \vspace{3mm}
\endminipage
\caption{MNIST: Latent geometry visualisation}
\label{abc}
\end{figure}

\begin{figure*}[!t]
\minipage{0.245\textwidth}
  \includegraphics[width = \textwidth]{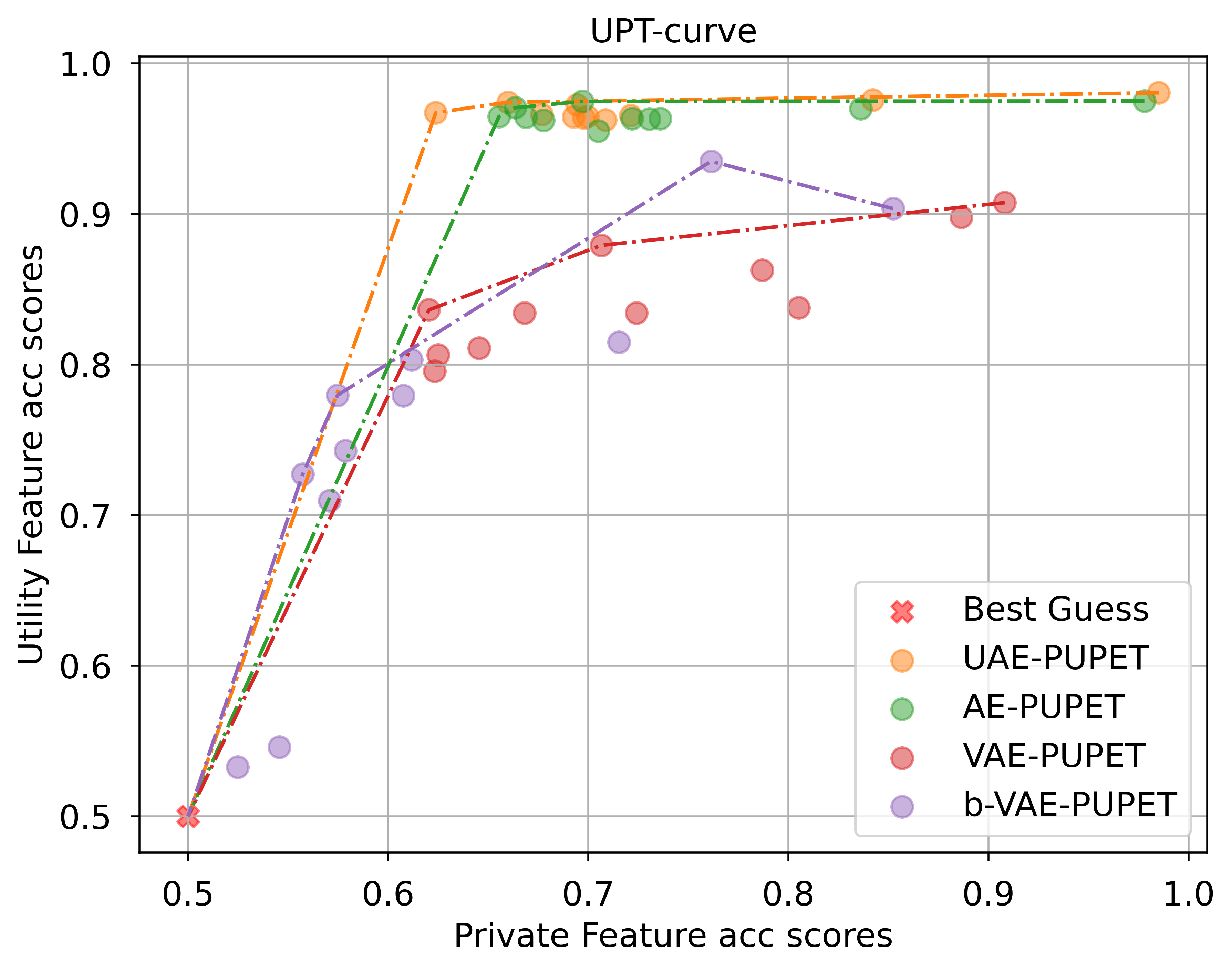}
  \subcaption{MNIST}
    \label{fig:upt-mnist}
\endminipage
\minipage{0.245\textwidth}
  \includegraphics[width = \textwidth]{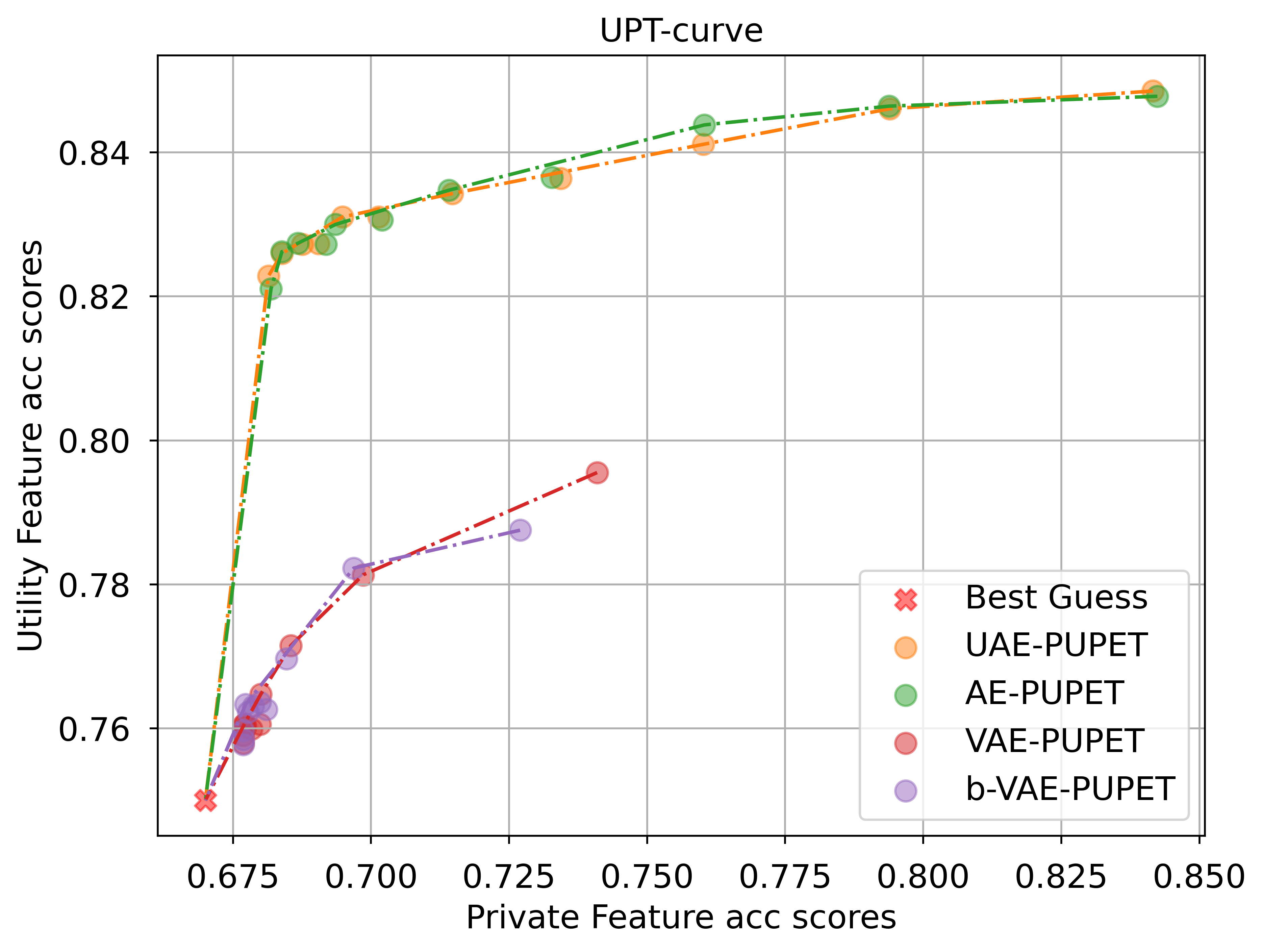}
  \subcaption{UCI Adult}
    \label{fig:upt-uci-adult}
\endminipage
\minipage{0.245\textwidth}
  \includegraphics[width = \textwidth]{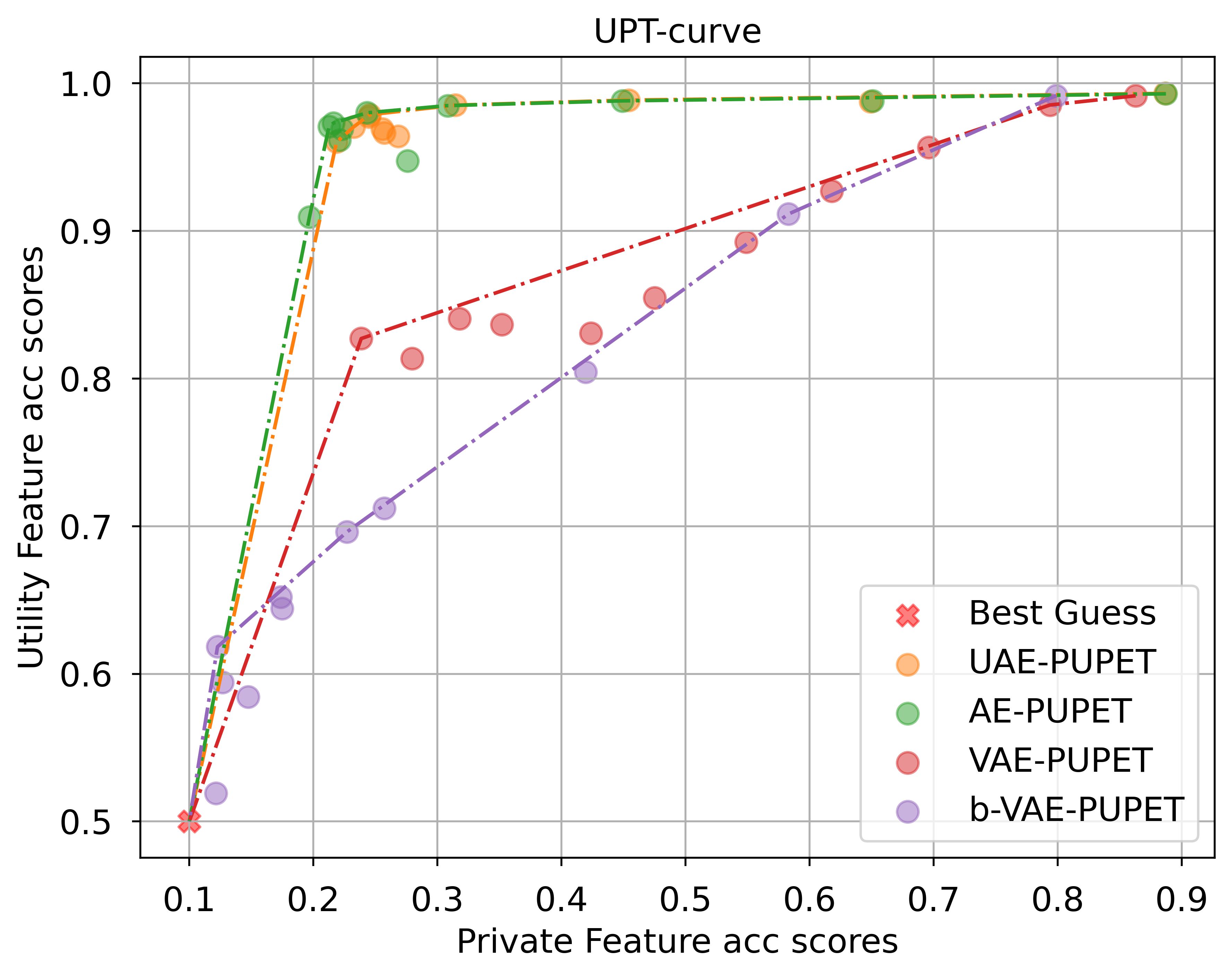}
  \subcaption{Fashion MNIST}
    \label{fig:upt-fashion-mnist}
\endminipage
\minipage{0.245\textwidth}
  \includegraphics[width = \textwidth]{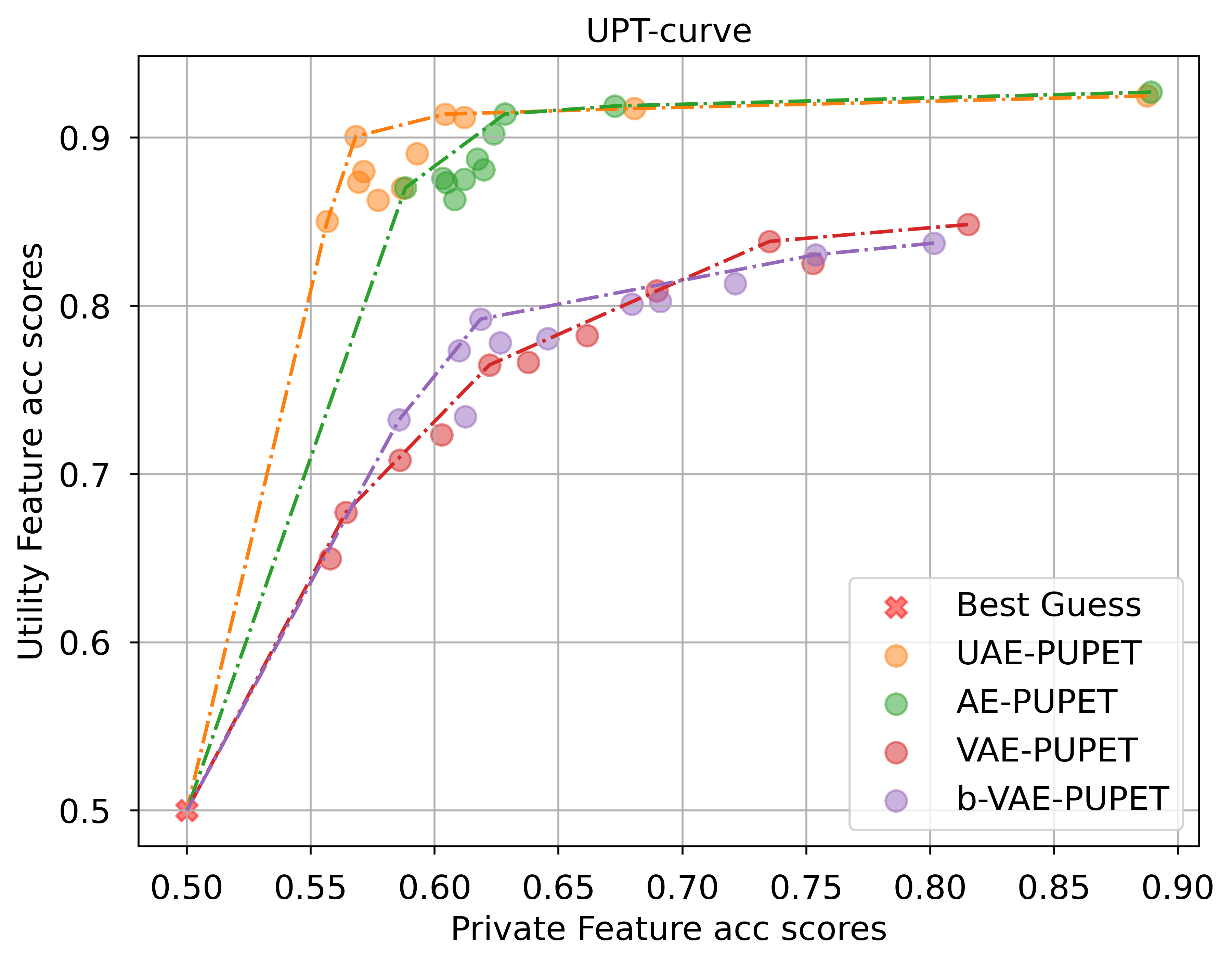}
  \subcaption{US Census}
    \label{fig:upt-us-census}
\endminipage
\caption{UPT Curves: Utility Privacy Tradeoff curves}
\label{fig:upt-curves}
\end{figure*}

In Fig. \ref{abc}, we provide a visualisation of latent variables before and after the application of our privacy mechanism when the dimension of the latent variable is two. The scatter plot does not have a closed structure because latent variables are not required to follow a Gaussian distribution. In Fig. \ref{fig:latent 1}, we can observe distinct regions in the two-dimensional space for the private feature -- even or odd, which makes the correct prediction of private feature possible for the adversary. However, after applying the privacy mechanism, it can be seen that the private feature -- even or odd -- doesn't have distinct regions in the two-dimensional space and the data points under a similar region sometimes correspond to even and sometimes to odd, making the correct prediction of the private feature difficult even for the \emph{best performing adversary}-- an adversary that classifies the private feature with the highest accuracy.

\begin{figure*}[!t]
\minipage{0.195\textwidth}
  \includegraphics[width = \textwidth]{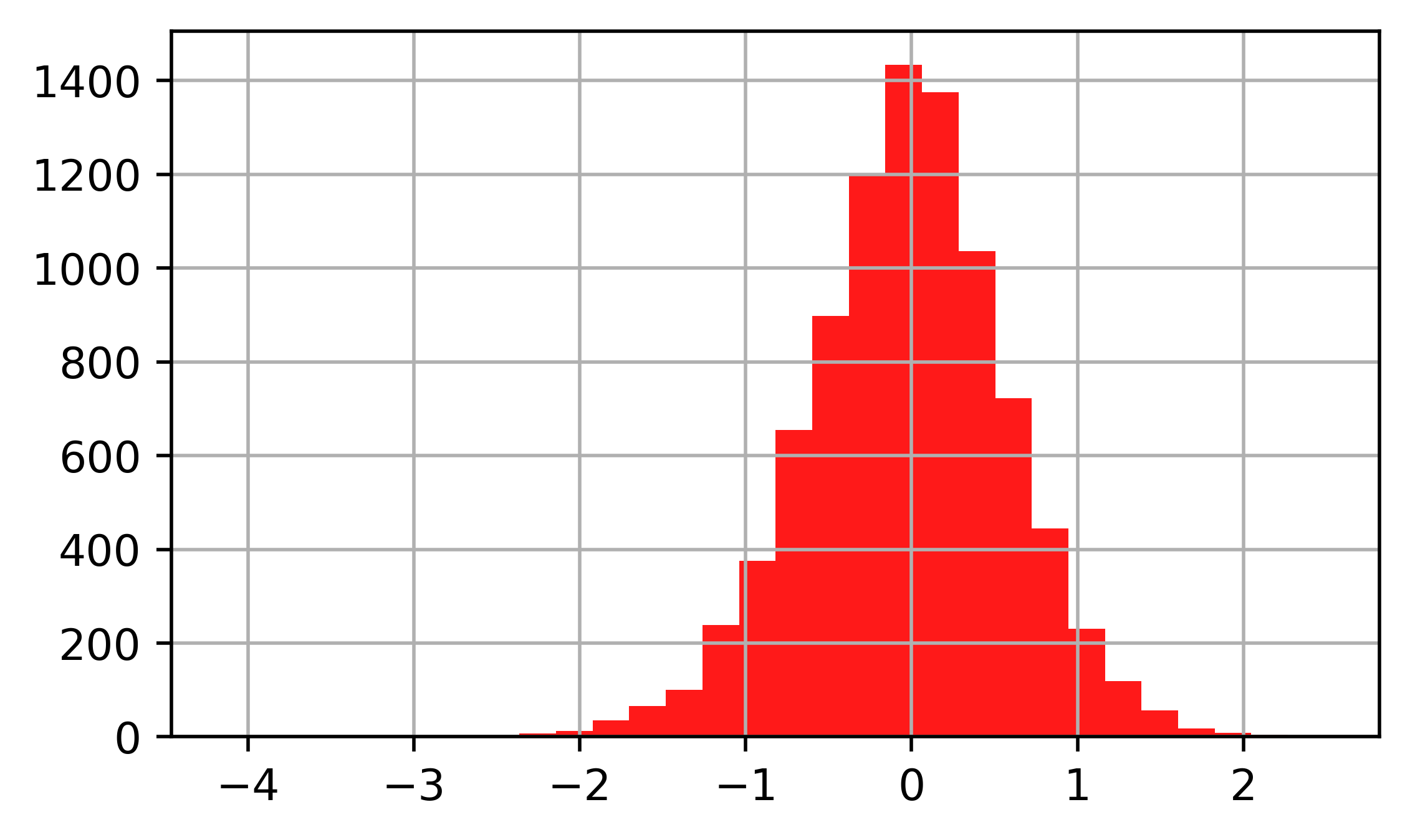}
    \label{fig:latent_a}
\endminipage
\minipage{0.195\textwidth}
  \includegraphics[width = \textwidth]{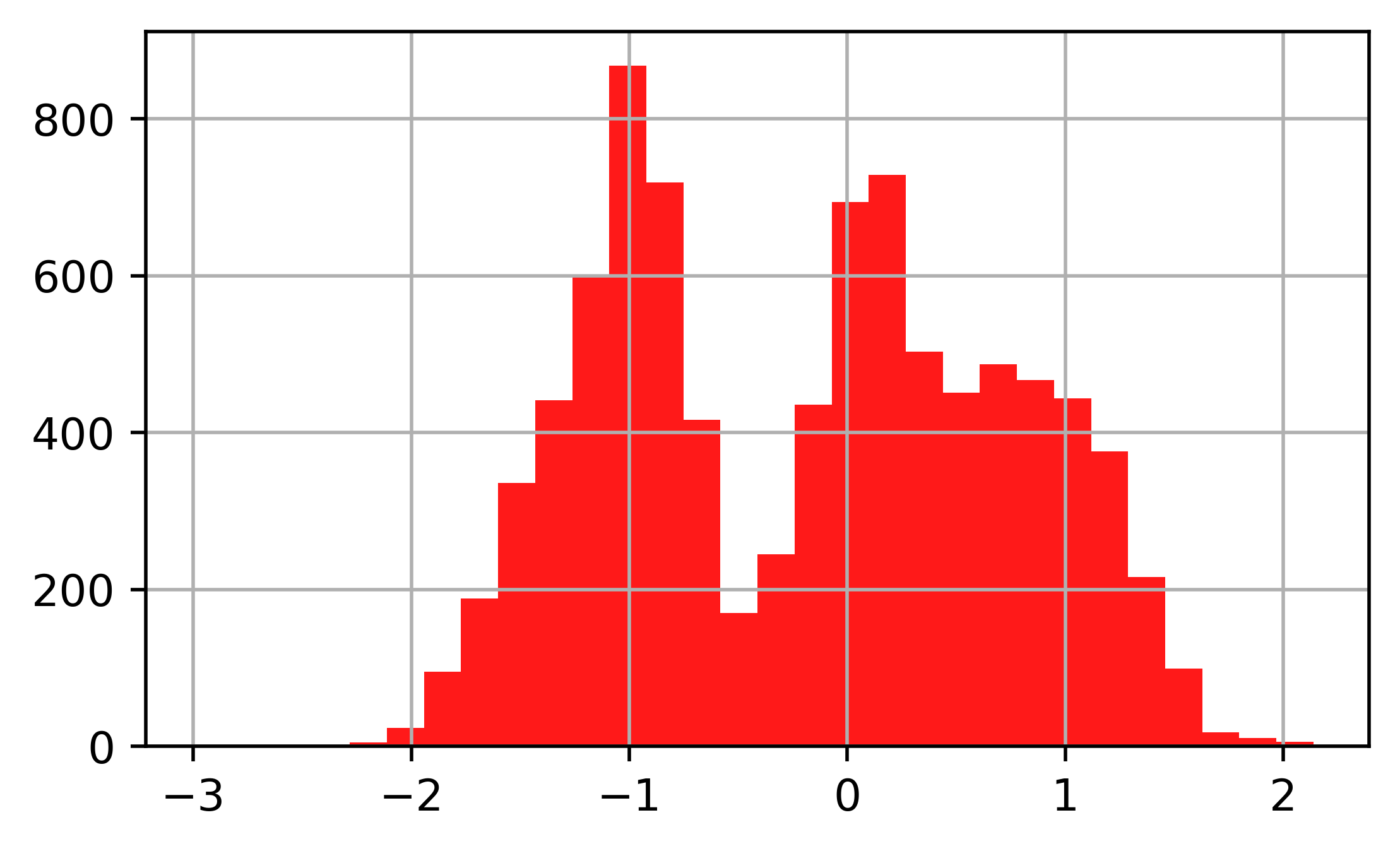}
    \label{fig:latent_b}
\endminipage
\minipage{0.195\textwidth}
  \includegraphics[width = \textwidth]{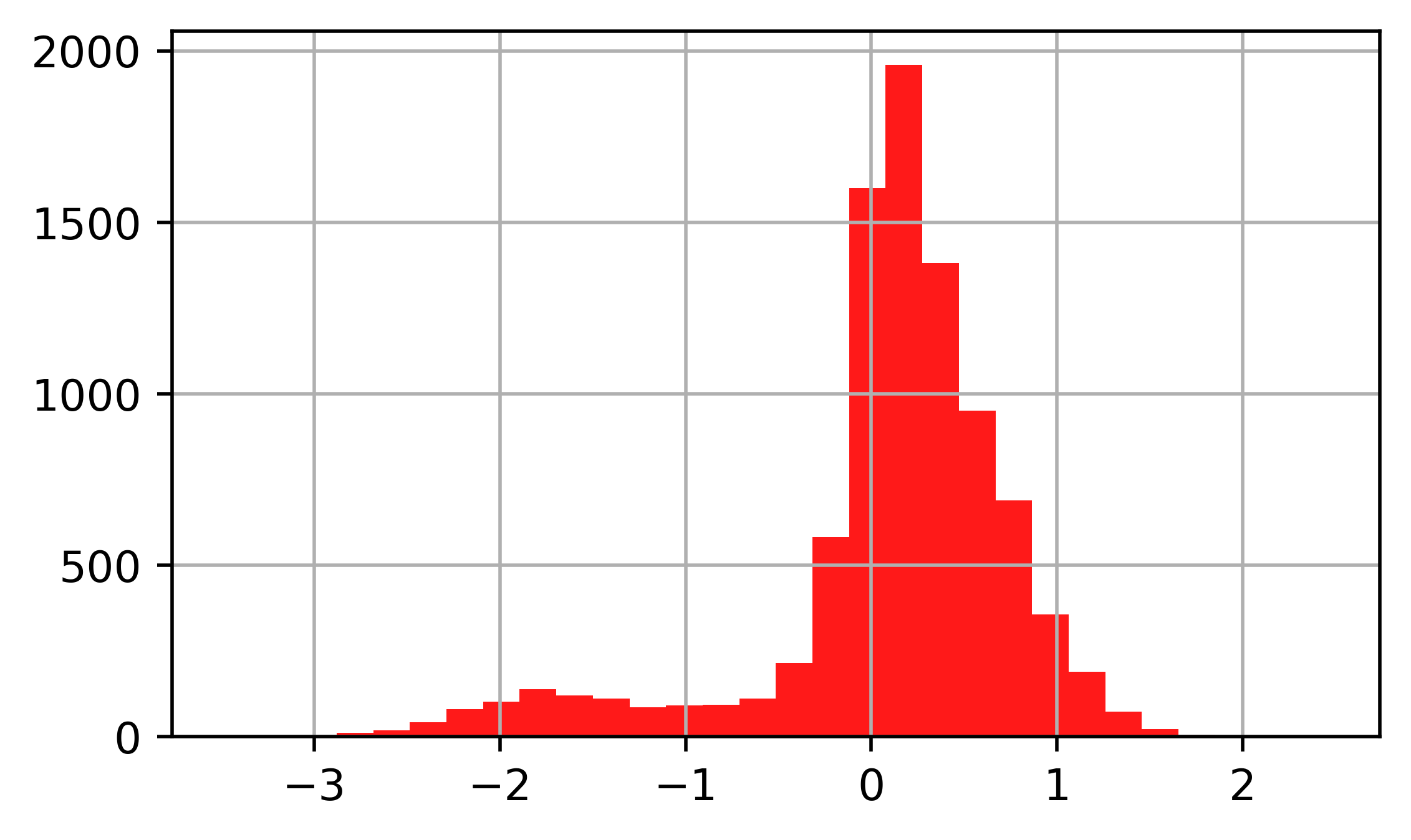}
    \label{fig:latent_c}
\endminipage
\minipage{0.195\textwidth}
  \includegraphics[width = \textwidth]{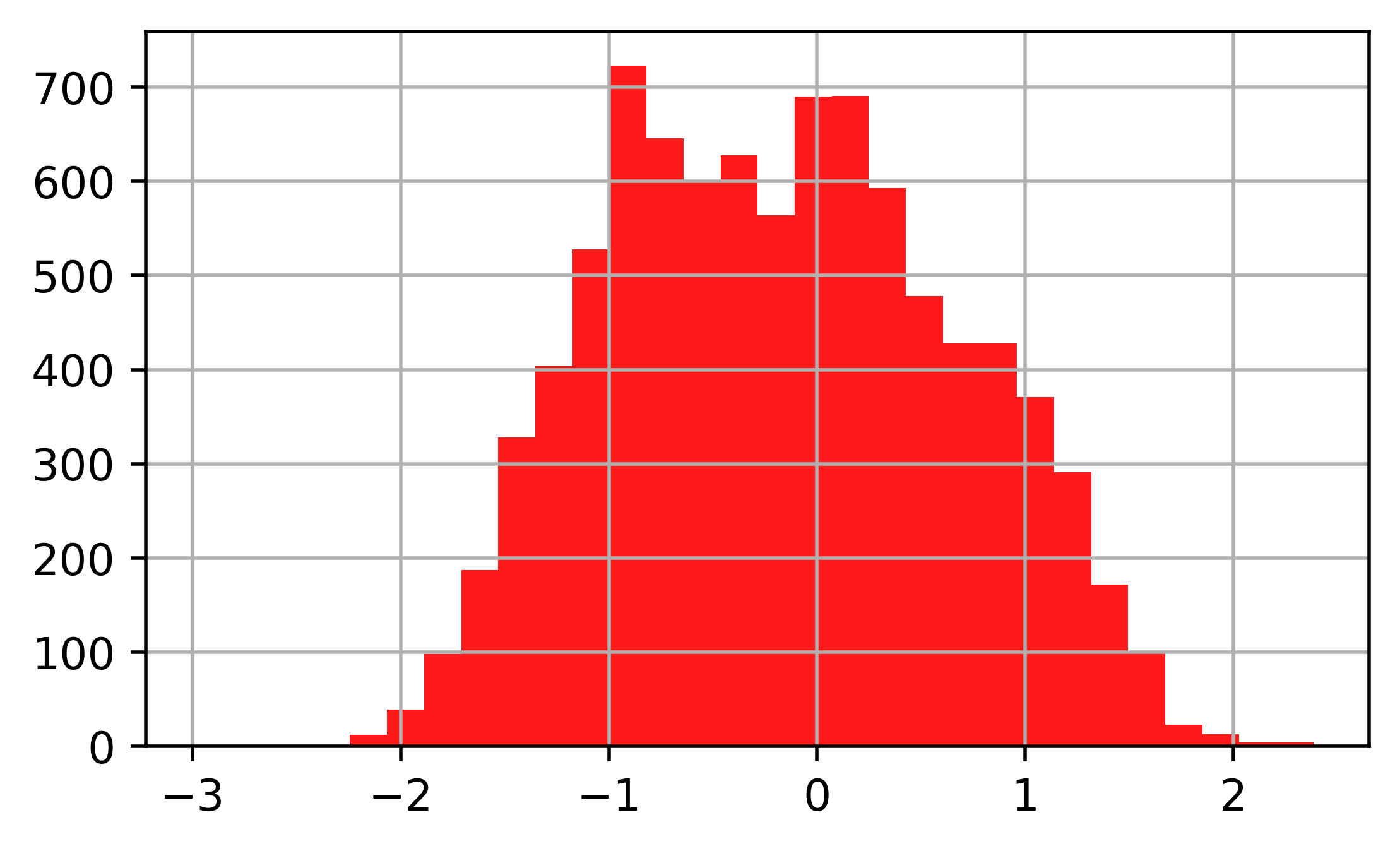}
    \label{fig:latent_d}
\endminipage
\minipage{0.195\textwidth}
  \includegraphics[width = \textwidth]{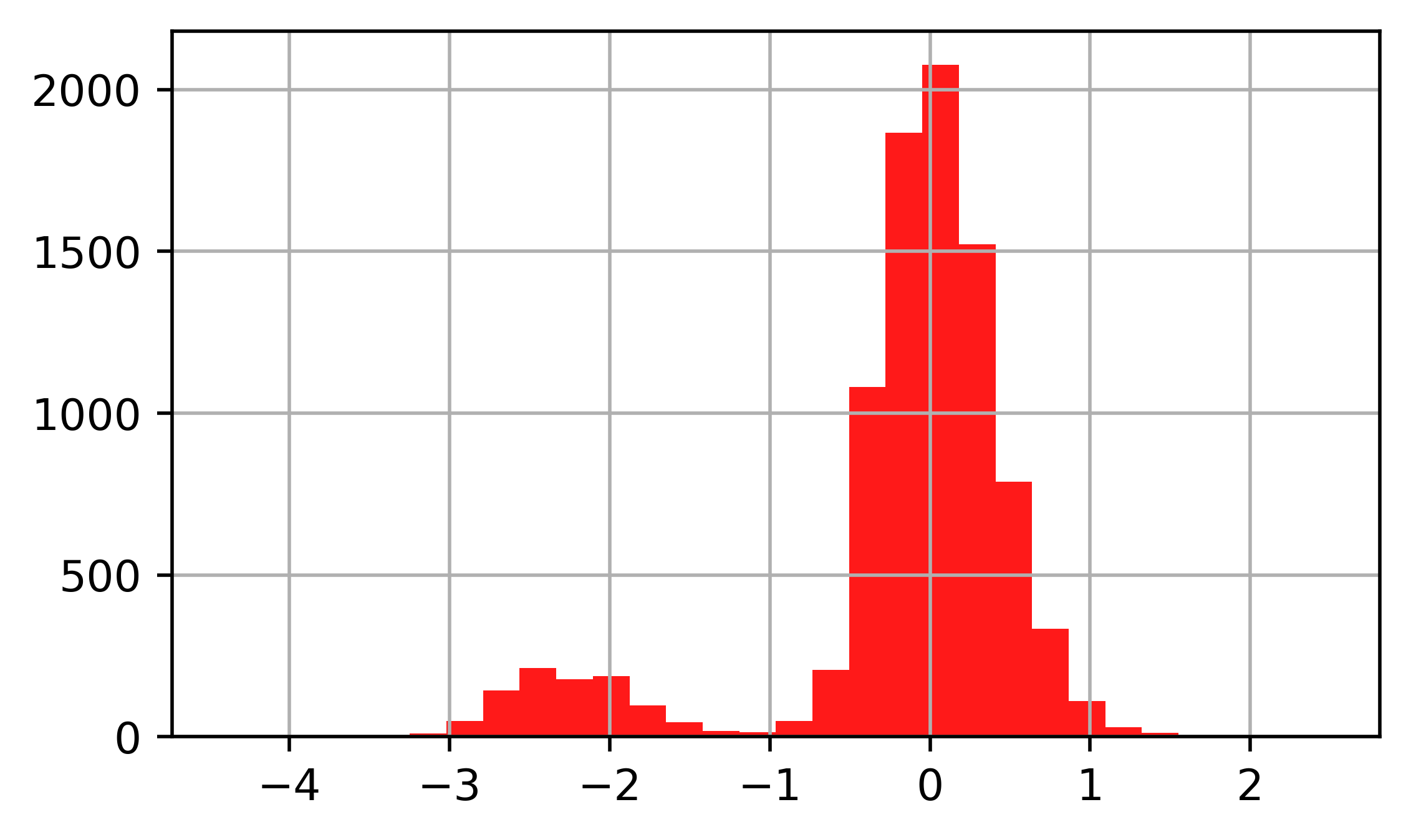}
    \label{fig:latent_e}
\endminipage
\caption{UCI Adult: Different distributions of distortions after using UAE-PUPET}
\label{fig:noise}
\end{figure*}

\subsection{UCI Adult}
In this experiment, we set our private feature as gender, and utility feature as income. Data pre-processing steps include removing all the data points with missing values, converting categorical variables to one-hot encoding representations and normalizing all the variables. We compare our privacy mechanism results to Variational Fair AutoEncoder (VFAE) \cite{louizos2017variational}, Lagrangian Mutual Information-based Fair Representations (LMIFR) \cite{DBLP:journals/corr/abs-1812-04218} and emb-g-filter \cite{chen2019distributed}. Similar to the MNIST experiment we also implement AE-PUPET, VAE-PUPET and $\beta$VAE-PUPET. It is evident through Table ~\ref{table-uci} that our proposed method UAE-PUPET has the least accuracy score of 0.6814, and AUROC score of 0.521 for the private feature and comparable accuracy score of 0.822, and AUROC score of 0.72 for the utility feature even after being tested under multiple adversaries. Fig. \ref{fig:upt-uci-adult} shows that UAE-PUPET and AE-PUPET perform quite similarly for this dataset. We should note that our implementation of AE-PUPET is almost identical to that in \cite{edwards2016censoring}, except their work considers alternative updates of gradients, unlike ours.

\begin{table}[!htb]
\resizebox{\columnwidth}{!}{
\begin{tabular}{|c|c|c|c|c|}
\hline
\multirow{2}{*}{Models} & \multicolumn{2}{c|}{Private attr.} 
&
\multicolumn{2}{c|}{Utility attr.} \\
\cline{2-5}
 & accuracy & AUROC & accuracy &  AUROC \\
\hline
without distortion(raw) & 0.83 & 0.75 & 0.84 & 0.76\\ 
VFAE \cite{louizos2017variational} & 0.802 & 0.703 & 0.851 & 0.761\\
LMFIR \cite{DBLP:journals/corr/abs-1812-04218} & 0.728 & 0.659 & 0.829 & 0.741\\
emb-g-filter \cite{chen2019distributed} & 0.717 & 0.632 & 0.822 & 0.731\\
VAE-PUPET & 0.698  & 0.573 & 0.781 & 0.597\\
$\beta$VAE-PUPET & 0.696  & 0.58 & 0.782 & 0.607\\
AE-PUPET &  0.6819 & 0.525 & 0.821  & 0.71 \\
\rowcolor{LightCyan}
UAE-PUPET & 0.6814  & 0.521 & 0.822 & 0.72\\
\hline
\end{tabular}
}
\caption{\small UCI Adult accuracy  and AUROC result comparison}
\label{table-uci}
\end{table}

Recall that our privacy mechanism does not add external noise during training. However, the regularization term provided by the discriminator losses when updating the $\gamma$ parameters reflect a distortion. By not forcing any specific noise distribution when training, the privacy mechanism has a higher degree of freedom to produce different distributions of distortion on its own. Fig. \ref{fig:noise} illustrates the most common distortion distributions observed on the UCI Adult dataset when privatizing using UAE-PUPET. The reason for the appearance of different distributions has to do with multiple statistical properties of the training data, which is an interesting area of research and will be the subject of future work.

\subsection{Additional Results}
\begin{figure}[!htb]
\centerline{\includegraphics[width=0.5\textwidth]{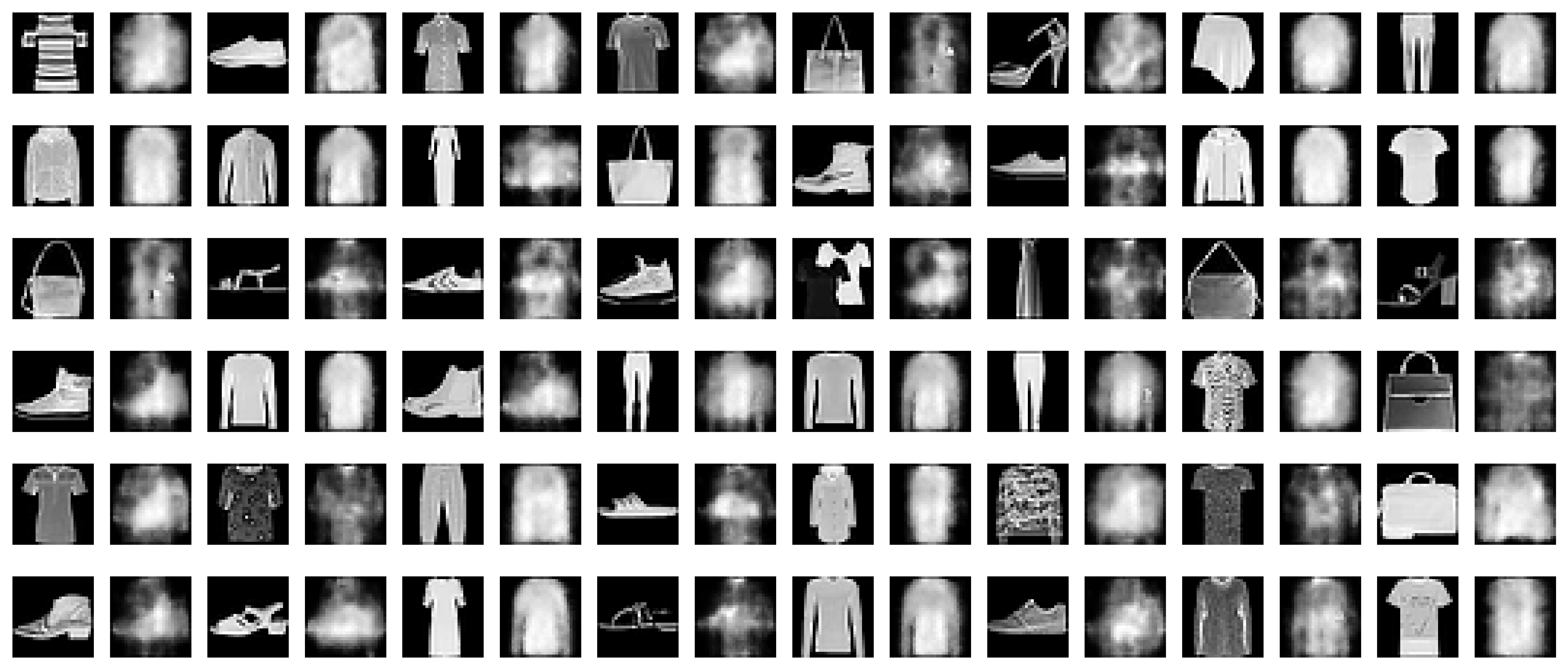}}
\caption{\textbf{Fashion MNIST}: Odd and even adjacent columns show original and privatized versions (UAE-PUPET) respectively.}
\label{fig3}
\end{figure}

Besides MNIST and UCI Adult, we also conduct experiments on Fashion MNIST and US Census Demographic Data. These experiments employ private and utility features that do not compare to any existing method; they are being presented solely for the purpose of demonstrating how different PUPET models worsen the adversaries' results. For Fashion MNIST dataset, the private feature is the identity of the fashion article (T-shirt/top, Trouser, Pullover, Dress, Coat, etc.) and the utility feature is encoded on two labels: Upper (meaning T-shirt/top, Pullover, Dress, Coat, Shirt) and Miscellaneous. Fig. \ref{fig3}, shows the privatized images, along with their original versions. We see that the privatized images appear to have been changed to different articles of clothing. We also notice blurriness of privatized images, sometimes appearing to be comprised of two different images juxtaposed on one. Our experiment shows a drop of inference accuracy on private feature from 98\% to 21\% using UAE-PUPET and AE-PUPET, 23\% using VAE-PUPET, 41\% using $\beta$VAE-PUPET under \emph{best performing adversary} whereas the inference accuracy on the utility feature decreases slightly from 99\% to 96\% using UAE-PUPET, 97\% using AE-PUPET, 82\% using VAE-PUPET and 80\% using $\beta$VAE-PUPET.

Similarly, US Census Demographic Data is the American Community Survey data for 2017 that consists of 74,001 records for different counties. It has a total of 37 features, out of which we select the sixteen features which are highly correlated to each other, similar to the setting in \cite{9500410}. Examples of some of the selected features include the men population, women population, population of citizens eligible to vote, per capita income, percentage of population unemployed etc. Among the sixteen features, we select Employed as the utility and Income as the private feature. Unlike \cite{9500410}, we further categorize the utility feature into two labels i.e. $\leq 2000$ or $> 2000$ and private feature into two labels i.e. $\leq 55000$ or $> 55000$ to make it a balanced classification problem. All the other fourteen features are continuous, and thus we normalize them based on their mean and standard deviation. Similarly, we ignore data points that have missing values. We use a total of 43,657 data points for training, and 29,105 data points for testing purposes. Our experiments show that the accuracy for private features drops down from 88\% to 56\% using UAE-PUPET, 58\% using AE-PUPET, 68\% using VAE-PUPET, 61\% using $\beta$VAE-PUPET, and the utility accuracy drops from 92\% to 90\% using UAE-PUPET, 87\% using AE-PUPET and 80\% using VAE-PUPET, and 79\% using $\beta$VAE-PUPET. This experiment again clearly shows better performance of UAE-PUPET over other models. Please refer to Fig. \ref{fig:upt-fashion-mnist} and Fig. \ref{fig:upt-us-census} for the complete comparison of these PUPET models.

\subsection{Data-Type-Aware Condition}
In this section, we discuss the data-type-aware condition and its result. We concentrate on the UCI Adult dataset since it is the only one out of four datasets used, that contains categorical variables. The exponential mechanism \cite{exponential} in differential privacy is designed specifically for discrete/categorical variables where classes are selected based on a scoring function. Similarly, \cite{10.5555/2525373.2525390} proposed a noise addition technique to the categorical data. However, these formulations solve different problem than ours. To the best of our knowledge, the existing literature on using an adversarial learning framework to solve the privacy-utility tradeoff problem does not discuss data-type-aware conditions which is to say, correctly encoding the categorical feature as one hot encoding that represents exactly one class, e.g.(a) (say [0, 0, 0, 1, 0]) after the distortion of data through the privacy mechanism. Instead, the privacy mechanism outputs different probability values for each class, e.g. (b) (say [0.1, 0.05, 0.03, 0.8, 0.02]) and this difference between (a) and (b) is referred to as noise \cite{edwards2016censoring, madras2018learning, chen2019distributed}. However, this notion of addition of noise for categorical features is heavily flawed as the user cannot specify a categorical variable as 10\% of the first class, 5\% of the second class, and so on. Therefore, we formulate data-type-aware conditions which use the precise representation of categorical variables before disclosing the data (refer to Algorithm \ref{alg:cap}).

Finally, we discuss about the experimental results for data-type-aware conditions on different versions of PUPET. Table ~\ref{table-uci2} shows that, even when enforcing constraints on the categorical variables, our PUPET models are capable of providing similar privacy and utility guarantees compared to the data-type-ignorant conditions. Additionally, we notice that UAE-PUPET and AE-PUPET results are again very similar in performance. With more experiments we observed that having small variance improves the performance slightly for this dataset which explains the similarity in results for UAE and AE based PUPETs -- recall that with zero variance, UAE's learning objective is same as AE's.

\begin{table}[!t]
\resizebox{\columnwidth}{!}{
\begin{tabular}{|c|c|c|c|c|}
\hline
\multirow{2}{*}{Models} & \multicolumn{2}{c|}{Private attr.} 
&
\multicolumn{2}{c|}{Utility attr.} \\
\cline{2-5}
 & accuracy & AUROC & accuracy &  AUROC \\
\hline
without distortion(raw) & 0.83 & 0.75 & 0.84 & 0.76\\ 
VAE-PUPET & 0.6843  & 0.5806 & 0.7831 & 0.6272\\
$\beta$VAE-PUPET & 0.6791  & 0.53 & 0.771 & 0.5813\\
AE-PUPET &  0.6832 & 0.5414 & 0.8245  & 0.728 \\
\rowcolor{LightCyan}
UAE-PUPET & 0.6817  & 0.5361 & 0.8220 & 0.7151\\
\hline
\end{tabular}
}
\caption{\small UCI Adult: data-type-aware conditions results}
\label{table-uci2}
\end{table}

\section{Conclusion}\label{conclusions}
In this paper, we introduced a novel UAE-based privacy mechanism (UAE-PUPET), and showed that it can attain better privacy-utility tradeoffs than the existing works. Additionally, the results for AE-PUPET were found to be very close to UAE-PUPET, but the use of VAE and $\beta$VAE showed poor results on all the datasets. UAE-PUPET is still better than AE-PUPET in the sense that it can behave like an AE-PUPET if required but the converse is not true. Better overall results with UAE-PUPET and AE-PUPET imply that forcing a Gaussian distribution on the latent variable of autoencoders (such as in VAE and $\beta$VAE) appears to hinder the privacy mechanism. Our privacy mechanism is focused on end-to-end stochastic mapping of input to the output. However, VAE and $\beta$VAE have a KL-divergence term in their learning objective, which  deviates from only focusing on the reconstruction term. The results for VAE and $\beta$VAE on UCI Adult is particularly poor because of the heterogeneous nature of the data. \cite{https://doi.org/10.48550/arxiv.2006.11941} formulate VAEs to handle the heterogeneous nature of data but this is a two stage process which adds complexity and is harder to incorporate with our settings. Instead, the use of UAE helps deal with heterogeneous data types, behaves like an AE if required, and can still add randomness in the latent space like VAE and $\beta$VAE.

\bibliographystyle{IEEEtran}
\bibliography{IEEEabrv, references}

\end{document}